\documentclass[letterpaper, 10 pt, conference]{ieeeconf}  

\IEEEoverridecommandlockouts                              

\usepackage{graphics} 
\usepackage{epsfig} 

\usepackage{caption}
\usepackage{subcaption}
\usepackage{float}
\usepackage{multirow}
\usepackage{colortbl,hhline}
\usepackage{amsmath} 
\usepackage{amssymb}  
\usepackage[fleqn,tbtags]{mathtools}
\usepackage{cancel}

\usepackage{color,soul}
\usepackage[bordercolor=white,backgroundcolor=gray!30,linecolor=black,colorinlistoftodos]{todonotes}

\usepackage[linesnumbered,ruled,vlined]{algorithm2e}
\usepackage[noend]{algpseudocode}
\usepackage{booktabs}
\usepackage{tabularx}
\usepackage{tabulary}
\usepackage{graphicx} 
\usepackage[export]{adjustbox}
\usepackage{cite}
\usepackage[super]{nth}
\usepackage{comment}
\let\labelindent\relax
\usepackage{enumitem}
\usepackage{framed}
\usepackage{makecell}

\newcommand{\ie}{{\em i.e.,~}}

\title{\LARGE \bf
ART-TEB: Adaptive Trajectory Planning for Mobile Robots \\ in Cluttered Environments
}

\author{Hahjin Lee and Young J. Kim
\thanks{The authors are with the Department of Computer Science and Engineering at Ewha Womans University in Korea
   ${\it \{hahjinlee|kimy\}@ewha.ac.kr}$.}%
}
\newcommand{\eg}{{\em e.g.,~}}

\DeclareMathOperator*{\argmin}{argmin}

\begin{document}

\maketitle
\thispagestyle{empty}
\pagestyle{empty}

\begin{abstract}

Trajectory planning for mobile robots remains a major challenge, particularly in cluttered environments, where existing planning algorithms often fail or produce suboptimal paths. To address this issue, we propose an adaptive trajectory refinement algorithm, ART-TEB, comprising two main stages. First, to ensure safety at the path-segment level, a segment-wise conservative collision test is applied, recursively subdividing risky path segments until the collision risk is eliminated. Second, to guarantee pose-level safety, pose correction based on separation direction and line search is applied, ensuring that each pose in the trajectory is collision-free and maximally clear from obstacles. Simulation results demonstrate that the ART-TEB achieves up to $3.12\times$ higher success rates and up to $23.7\times$ faster average planning times than state-of-the-art approaches. Furthermore, real-world experiments confirm that the robot can safely pass through highly constrained environments while maintaining rapid planning performance.

\end{abstract}

\section{Introduction} \label{section:intro}
Autonomous navigation of mobile robots has become a core technology in various domains such as logistics, industrial automation, and service robotics. In particular, real-time path planning that generates kinodynamically constrained trajectories is indispensable for robots to navigate safely and efficiently in cluttered environments. Local planners generally play a central role in generating such trajectories subject to collision and kinodynamic constraints. 
Among the existing local planners, the dynamic window approach (DWA) \cite{fox1997dynamic}, the elastic band (EB) \cite{quinlan1993elastic}, and the timed-elastic-band (TEB) \cite{rosmann2012trajectory,rosmann2013efficient,rosmann2017integrated} are widely used in both academia and industry thanks to their practicality. However, DWA is limited by its short planning horizon, whereas EB struggles to ensure kinodynamic feasibility. TEB can generate long-horizon, time-optimal trajectories with guaranteed kinodynamic feasibility by employing spatio-temporal optimization. These advantages of TEB have contributed to its widespread adoption across diverse domains, such as autonomous scanning of indoor environments \cite{lee2025scanning}, diffusion-based global path planning frameworks \cite{liu2024dipper}, and autonomous robot exploration in unknown environments \cite{han2022autoexplorer}.

TEB computes a time-optimal trajectory $\mathcal{T}^{*}$ that satisfies various constraints on the robot. 
$\mathcal{T}^{*}$ is obtained through iterative, penalty-based optimization with weights $\alpha_k$ assigned to each cost term $f_k$ including the temporal resolution $\Delta t$:
\begin{equation}
    \mathcal{T}^{*} = \argmin_{\mathcal{T}} \sum_{k} \alpha_k f_k(\mathcal{T}(\Delta t))
\end{equation}
However, TEB-based trajectory planning remains a significant challenge in constrained environments. 
The optimization of the TEB may guide the path to penetrating obstacles \cite{zhang2023improve}.
Also, TEB tends to generate sparse waypoints near obstacles, likely increasing the collision probability in cluttered environments.
Finally, TEB also suffers from inefficiencies in unconstrained space, as it unnecessarily employs a high temporal resolution even in free space, thereby increasing both optimization and planning costs. As a result, in our extensive experiments in Sec.~\ref{sec:exp}, TEB fails to plan $16\% \sim 41\%$ of the well-known test cases.

{To ensure the reliability of TEB in such challenging environments, robust collision resolution is required. Advanced collision resolution methods, such as continuous collision detection (CCD) and penetration depth, provide rigorous guarantees for collision resolution. However, these methods are rarely employed in real-time robot navigation due to the computational overhead of geometric evaluations \cite{lin2017collision}, \cite{jaitly2025analytic}.}

Our new method, ART-TEB, introduces an efficient adaptive trajectory-refinement algorithm that employs robust collision resolution with minimal overhead, suitable for online navigation. ART-TEB employs a two-level collision-resolution strategy operating at both the {\em continuous} segment and the {\em discrete} pose levels. At the path-segment level, Lipschitz-based CCD verifies the collision risk of each segment using conservative bounds. For this collision checking, we employ a circular footprint, a common standard in mobile robot navigation \cite{siegwart2011introduction}. Segments identified as risky are selectively subdivided by inserting additional waypoints, which are subsequently corrected using the pose-level method. At the pose level, collisions are resolved at low computational cost through a simple separation direction (SD) calculation followed by a line search. Together, these algorithms allow the trajectory to achieve reliable collision avoidance with finer resolution in constrained space while remaining sparse in free space, thereby achieving adaptive refinement.  
Integrating SD- and CCD-based collision-resolution methods into a time-parameterized local planner, ART-TEB iteratively adjusts temporal resolution and orientation, restoring kinodynamic feasibility and providing robust initialization, guiding subsequent optimization into a collision-free solution.

To evaluate our algorithm, we conducted experiments in various simulation and real-world scenarios. ART-TEB achieves up to $1.14\times \sim3.12\times$ higher success rate while reducing average planning time by $1.44\times\sim23.7\times$ compared to the recent, optimization-based methods such as TEB\cite{rosmann2017integrated} and egoTEB\cite{ smith2020egoteb}, as well as DWA\cite{fox1997dynamic} and MPC planners\cite{ rosmann2021online}. In addition, ART-TEB achieved second place in both simulation and real-world phases of the {\em 2026 BARN Challenge}, an international competition that aims to solve robotic navigation in highly constrained environments. 

\section{Related Works}
This section briefly reviews previous work on local planning algorithms, TEB-based approaches, and advanced collision-detection methods for local planning.

\subsection{Local Planning Algorithms}

Traditionally, reactive planning algorithms select actions from the immediate environmental context, including methods that employ potential fields \cite{khatib1986real} and velocity-obstacle formulations \cite{fiorini1998motion}. In particular, the DWA \cite{fox1997dynamic} is a practical, widely used method based on a sampling-based predictive control algorithm. Optimization-based planners \cite{quinlan1993elastic, ratliff2009chomp, kalakrishnan2011stomp} can refine trajectories into smoother and more feasible paths by explicitly considering motion constraints, while optimal-control-based methods integrate dynamics directly into trajectory generation based on MPC \cite{ rosmann2021online}. More recently, learning-based planners have emerged, leveraging neural networks including transformers \cite{damanik2024lics}, diffusion-based models \cite{yu2024ldp}, and reinforcement learning strategies \cite{mohammad2025soft}.

\subsection{Timed Elastic Band}

TEB is a local trajectory planning method that computes a time-optimal path while considering the robot's kinodynamic constraints and collision avoidance requirements \cite{rosmann2012trajectory,rosmann2013efficient,rosmann2017integrated}. To improve the optimization efficiency and collision avoidance of TEB, \cite{smith2020egoteb} resolves the mismatch between occupancy grids and factor graph representations by using an egocentric perception space. \cite{xi2024safe} introduces a dynamic global point adjustment module that enables the robot to follow the central path of the free space. \cite{11128768} introduces online parameter adaptation to adjust safe margins and planner behavior in real time. However, as these approaches rely on TEB optimization, they cannot handle trajectories that penetrate obstacles. To address this issue, \cite{zhang2023improve} introduced an obstacle gradient to prevent trajectories from being trapped in obstacles. In addition, \cite{zhang2025ga} focused on collision avoidance by grouping obstacles into convex clusters and extending the local goal into lines. However, since these methods search multiple paths through different homotopy classes, they incur significant planning-time overhead. On the other hand, our ART-TEB algorithm eliminates collisions through direct collision resolution and adaptive path bisection, thereby reducing total planning time and the number of failures.

\subsection{Collision Detection for Local Planning}

Collision detection and proximity computation have been extensively studied in the literature \cite{lin2017collision}.
Advanced collision-detection techniques, such as CCD, has been integrated into local planning. One of the early efforts, \cite{redon2005practical} employed CCD to obtain precise contact information to efficiently sample contact space. \cite{pan2012collision} employed cubic B-spline trajectories combined with CCD to generate dynamically feasible paths. More recently, \cite{merkt2019continuous} introduced a continuous-time collision avoidance term into trajectory optimization.

{Meanwhile, to resolve infeasible configurations, local planners often require geometric collision resolution, such as penetration depth (PD) computation. For instance, retraction-based motion planning employs penetration depth to pull collided configurations toward the collision-free region \cite{zhang2008efficient}, but these computations incur significant overhead \cite{zhang2007fast}. Thus, many methods rely on sampling of the boundary of configuration-space obstacles \cite{lee2012sr}. Although recent optimization frameworks incorporate distances between geometries directly into solvers as collision-avoidance constraints \cite{jaitly2025analytic,helling2023dual}, these approaches substantially increase the computational burden.} 

While these methods provide accurate collision information or correction cues, they are difficult to directly apply to mobile local navigation, requiring high-frequency planning. Our approach instead reformulates CCD and PD into computationally efficient approximations for rapid, real-time trajectory refinement in mobile navigation.


\section{Adaptive Trajectory Refinement Algorithm}

\subsection{Problems in TEB-based Planning}

\begin{figure}[htb!]
    \centering
    \includegraphics[width=\columnwidth]{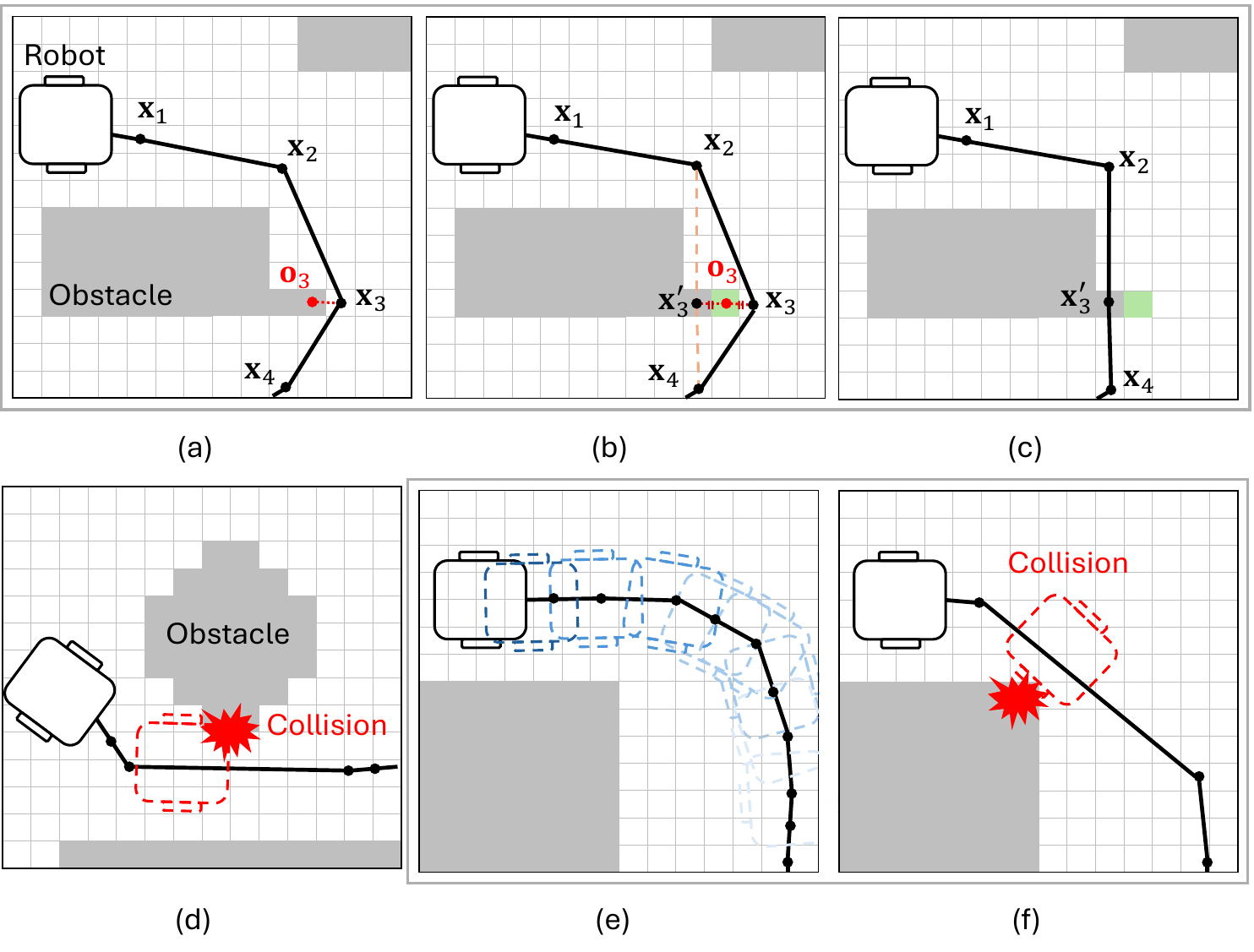}
    \caption{{\bf Limitations of TEB} (a) $\mathbf{o}_3$ is the closest to $\mathbf{x}_3$. The trajectory can penetrate into obstacles since (b) $\lVert \mathbf{x}_3-\mathbf{o}_3\rVert = \lVert \mathbf{x}_3' - \mathbf{o}_3\rVert$ and (c) $\mathbf{x}_3'$ yields a shorter trajectory than $\mathbf{x}_3$. (d) Waypoints are sparsely placed near obstacles, leading to collisions. (e) The original temporal resolution in the trajectory. (f) Reducing the resolution can cause collisions.}
    \label{fig:problem}
\end{figure}

\begin{figure*}[ht!]
{\includegraphics[width=\textwidth]{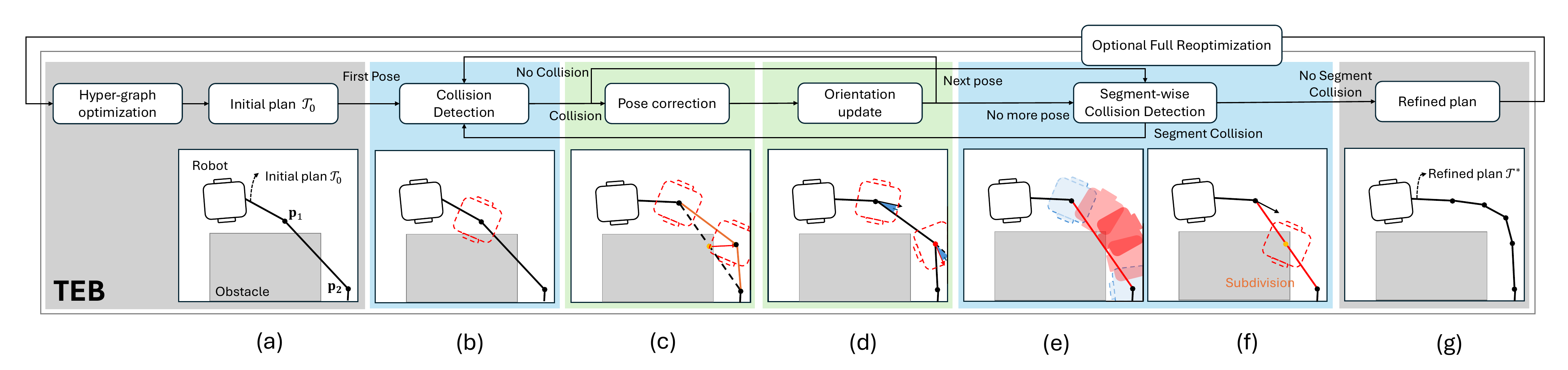}}
\captionof{figure}{ \textbf{Algorithm Overview.} 
(a) An initial plan from TEB hyper-graph optimization. (b–d) Iterative collision detection, pose correction, and orientation updates ensure all poses are collision-free. (e-f) Segment-wise CCD subdivides risky segments, with new poses refined through (b-d). (g) A collision-free trajectory is obtained. This trajectory is then fed back to the hyper-graph optimization, forming an iterative planning pipeline.
}
\label{fig:whole_pipeline_new}
\end{figure*}

TEB-based path planning in cluttered environments remains a significant challenge, as illustrated in Fig.~\ref{fig:problem}.
First, each waypoint $\mathbf{x}_i$ in TEB considers the closest point on the obstacle (\eg $\mathbf{o}_3$ is the nearest point for $\mathbf{x}_3$ in Fig.~\ref{fig:problem}(a)). 
Since $\lVert \mathbf{x}_3-\mathbf{o}_3\rVert = \lVert \mathbf{x}_3' - \mathbf{o}_3\rVert$ in Fig.~\ref{fig:problem}(b), $\mathbf{x}_3'$ has the same distance cost as $\mathbf{x}_3$ but 
the trajectory including $\mathbf{x}_3'$ has a shorter path length than the one including $\mathbf{x}_3$. As a result, the optimizer may select the path that passes through $\mathbf{x}_3'$, yielding collisions, as illustrated in Fig.~\ref{fig:problem}(c) \cite{zhang2023improve}. In contrast, our method directly relocates waypoints (\eg $\mathbf{x'}_3$) to become maximally clear from the obstacles (detailed in Sec.~\ref{subsection:pose_corr}).

Secondly, TEB tends to generate unnecessarily dense waypoints in free space while becoming sparse near obstacles, likely increasing the collision probability in narrow passages as illustrated in Fig.~\ref{fig:problem}(d).
Finally, since TEB maintains a high temporal resolution even in unconstrained space, as shown in Fig.~\ref{fig:problem}(e), it may unnecessarily increase the optimization and planning time. However, simply lowering the resolution is not viable, as sparse waypoints may induce the collision in less constrained regions, as illustrated in Fig.~\ref{fig:problem}(f). 
We address these issues by adaptively distributing the temporal resolution of waypoints, keeping it lower in free space while increasing it near obstacles (detailed in Sec.~\ref{subsection:ccd}).

\subsection{Overview} \label{subsection:overview}
In our algorithm, the planned trajectory is composed of piecewise time-parameterized curve segments, defined by a tuple $\mathcal{T} = (\chi, \tau)$, where $\chi = \{\mathbf{p}_i \in \mathrm{SE}(2) \mid 0\le i \le n \}$ is a finite sequence of robot poses (\ie waypoints) and $\tau = \{\Delta t_i \in \mathbb{R}^+ \mid 0\le i \le n-1\}$ is a sequence of time intervals between consecutive poses $\mathbf{p}_i, \mathbf{p}_{i+1}$. 
We also discretize the planning environment as a 2D grid map $\mathcal{M}$, which consists of discrete grid cells $c$, being in occupied $c \in \mathcal{O}$ or free space $c \in \mathcal{F}$. 
Then, our adaptive trajectory refinement algorithm proceeds iteratively as follows (also illustrated in Fig.~\ref{fig:whole_pipeline_new}):
\begin{enumerate}
\item We begin with an initial trajectory $\mathcal{T}_{0}=(\chi_0, \tau_0)$ provided by TEB with a coarse temporal resolution of $\tau_0$ to reduce the computational overhead of initial planning for the TEB (Fig.~\ref{fig:whole_pipeline_new}(a)). We refine the temporal resolution by identifying collision-risky regions based on the segment-wise CCD test (step 3) while keeping the resolution sparse for safe regions.

\item For the $k$-th iteration, each pose $\mathbf{p}_i \in {\chi}_{k}$ is examined for collisions 
(Fig.~\ref{fig:whole_pipeline_new}(b)). Poses in collision are immediately resolved through a pose correction step (Fig.~\ref{fig:whole_pipeline_new}(c) and Sec.~\ref{subsection:pose_corr}), followed by an orientation update (Fig.~\ref{fig:whole_pipeline_new}(d) and Sec.~\ref{subsection:ori_align}). 

\item Once all discrete poses $\forall \mathbf{p}_i \in \chi_k$ are determined to be collision-free, the path segment $S_i = \{\mathbf{p}_{i}, \mathbf{p}_{i+1}\}$, which connects the end poses $\mathbf{p}_{i}$ and $\mathbf{p}_{i+1}$, is examined for collision using segment-wise CCD (Sec.~\ref{subsection:ccd}). If the path segment $S_i$ is identified as risky (Fig.~\ref{fig:whole_pipeline_new}(e)), it is subdivided into $\{\mathbf{p}_{i}, \mathbf{p}_{i+\frac{1}{2}}, \mathbf{p}_{i+1} \}$ inserting intermediate poses $\mathbf{p}_{i+\frac{1}{2}}$ (Fig.~\ref{fig:whole_pipeline_new}(f)). 
For the inserted poses, collision detection is performed, and if any of them are found to be colliding, the correction procedure (step 2) is applied again to resolve the collision. Subsequently, the original $\Delta t_i$, corresponding to the time interval between $\mathbf{p}_i$ and $\mathbf{p}_{i+1}$, is also subdivided between the two new sub-segments in proportion to their motion displacements. 

\item The cycle (step 2 and 3) continues until all path segments $\forall i, S_i \subset \chi_*$ are confirmed to be collision-free, and $\mathcal{T}^{*}$ is produced as the final trajectory (Fig.~\ref{fig:whole_pipeline_new}(g)).

\item If $\mathcal{T}^{*}$ needs to be fully re-optimized for the constraints, step 1 can be fed with $\mathcal{T}_0=\mathcal{T}^{*}$ and reiterated until an optimization budget expires. 
\end{enumerate}

\subsection{Segment-wise CCD} \label{subsection:ccd}

\begin{algorithm}[t]
\setlength{\lineskip}{5pt}
\caption{Segment-wise CCD}
\label{alg:ccd_refinement}
\KwIn{Trajectory with collision-free poses $\chi =\{\mathbf{p}_0, \mathbf{p}_1, \dots, \mathbf{p}_n\}$ and time intervals $\tau = \{\Delta t_0, \Delta t_1, \dots, \Delta t_{n-1}\}$} \nllabel{ln:init}
Initialize priority queue $\mathcal{Q} \gets \emptyset$ \; \nllabel{ln:initq}
\For{each segment ${S}_i = \{\mathbf{p}_i, \mathbf{p}_{i+1}\} \subset \chi$}{\nllabel{ln:seg0}
    Compute $L_i = L(\mathbf{p}_i, \mathbf{p}_{i+1})$ \;\nllabel{ln:seg1}
    Push $({S}_i, \Delta t_i, L_i)$ into $\mathcal{Q}$ \;\nllabel{ln:seg2}
}
\While{$\mathcal{Q}$ is not empty}{ \nllabel{ln:while}
    Pop $({S}_i, \Delta t_i, L_i)$ \; \nllabel{ln:pop}
    \eIf{$L_i < d(\mathbf{p}_i) + d(\mathbf{p}_{i+1})$}{ \nllabel{ln:if}
        Accept $S_i$ as collision-free\; \nllabel{ln:free}
    }{
        $\mathbf{p}_{i+\frac{1}{2}} \gets \text{Bisect}({S}_i)$ \ ;\nllabel{ln:bisect}
        $S_i^{(1)} = \{\mathbf{p}_{i}, \mathbf{p}_{i+\tfrac{1}{2}}\} , S_i^{(2)} = \{\mathbf{p}_{i+\tfrac{1}{2}}, \mathbf{p}_{i+1}\}$ \; \nllabel{ln:segment}
        \If{$d(\mathbf{p}_{i+\frac{1}{2}}) \leq R$}{ \nllabel{ln:col}
            $\text{poseCorrection}(\mathbf{p}_{i+\frac{1}{2}})$ \; \nllabel{ln:pose_corr}
        }
        updateOrientations($\mathbf{p}_i, \mathbf{p}_{i+\frac{1}{2}}, \mathbf{p}_{i+1}$) \;\nllabel{ln:ori}
        $L_i^{(1)} = L(\mathbf{p}_i, \mathbf{p}_{i+\frac{1}{2}}), L_i^{(2)} = L(\mathbf{p}_{i+\frac{1}{2}}, \mathbf{p}_{i+1})$ \; \nllabel{ln:len}
      
        \(\begin{aligned}[t]
        \Delta t_{i}^{(1)} &\gets \Delta t_i \frac{L_i^{(1)}}{L_i^{(1)} + L_i^{(2)}} \\
        \Delta t_{i}^{(2)} &\gets \Delta t_i \frac{L_i^{(2)}}{L_i^{(1)} + L_i^{(2)}} \;\;\;\;\;\;\; ; 
        \end{aligned}\)\; \nllabel{ln:time}
        
        Push $(S_i^{(1)}, \Delta t_{i}^{(1)}, L_i^{(1)})$ into $\mathcal{Q}$ \;\nllabel{ln:q1}
        Push $(S_i^{(2)}, \Delta t_{i}^{(2)}, L_i^{(2)})$ into $\mathcal{Q}$\;\nllabel{ln:q2}

        $\chi \gets \chi \cup \{\mathbf{p}_{i+\tfrac{1}{2}}\}$\; \nllabel{ln:pose}
        $\tau \gets (\tau \setminus \{\Delta t_i\}) \cup \{\Delta t_i^{(1)}, \Delta t_i^{(2)}\}$ \;\nllabel{ln:inter}
    }
}
\KwOut{$\mathcal{T}^* = (\chi, \tau)$}
\end{algorithm}

Assuming that we have a trajectory only consisting of collision-free poses, we proceed to check for collisions for each path segment $S_i = \{\mathbf{p}_i, \mathbf{p}_{i+1}\} \subset \chi$ along the entire trajectory $\mathcal{T}=(\chi, \tau)$.
This problem is known as CCD for a trajectory, as opposed to conventional, discrete collision detection (DCD) for a discrete pose \cite{lavalle2006planning}. 
CCD typically calculates the swept volume or solves nonlinear equations to find the exact time of contact (ToC), which is the first time that two moving geometries come into contact. These operations are computationally expensive because they require geometric reasoning over the motion interval. To reduce this cost, we employ a Lipschitz-based condition that avoids the explicit computation of ToC. Precisely, our method conservatively verifies whether a trajectory segment is guaranteed to remain collision-free at low cost.

A path segment $S_i$ is guaranteed to be collision-free when a conservative upper bound for the motion displacement $L(\mathbf{p}_i, \mathbf{p}_{i+1})$ within the segment is less than the sum of the clearances at the two end poses $\mathbf{p}_i, \mathbf{p}_{i+1}$. {For efficient collision checking, the robot footprint is conservatively approximated by its circumscribed circle with radius $R$.} Accordingly, the bound can be expressed as $L(\mathbf{p}_i, \mathbf{p}_{i+1}) \equiv \Delta d + R\Delta\theta$ where $\Delta d$ and $\Delta \theta$ denote the position and orientation differences between $\mathbf{p}_i$ and $\mathbf{p}_{i+1}$, respectively. Then, a sufficient condition (\ie the CCD test) for certifying that $S_i$ does not contain a collision is given by
\begin{equation}\label{eq:ccd}
L(\mathbf{p}_i, \mathbf{p}_{i+1}) < d(\mathbf{p}_i) + d(\mathbf{p}_{i+1}), 
\end{equation}
where $d(\mathbf{p})$ corresponds to the minimum Euclidean distance between the obstacle and the robot footprint transformed by the current pose \(\mathbf{p}\) \cite{schwarzer2004exact}. 

In fact, the CCD condition in Eq.~\ref{eq:ccd} is a sufficient certificate derived from Lipschitz continuity as follows. 
The map $\mathbf{p}\mapsto d(\mathbf{p})$ is 1-Lipschitz with respect to the configuration-space metric whose path length is measured by $L$. By Lipschitz continuity with a lower-envelope bound, any interpolation $\gamma$ between $\mathbf{p}_i$ and $\mathbf{p}_{i+1}$ with $L$-length $l$ satisfies
\begin{equation}\label{eq:lcond3}
\min_{s\in [0,1]} d(\gamma(s)) \ge \frac{d(\mathbf{p}_i)+d(\mathbf{p}_{i+1})-\ell(\gamma)}{2}.
\end{equation}
Thus, if $\ell(\gamma)<d(\mathbf{p}_i)+d(\mathbf{p}_{i+1})$, the whole segment is collision-free. Our CCD test (Eq.~\ref{eq:ccd}) is exactly this certificate with $\ell(\gamma)$ replaced by the computable bound $L(\mathbf{p}_i,\mathbf{p}_{i+1})$.
Lipschitz-based bounds have also been used in \cite{aablad2021continuous}, where Lipschitz continuity is exploited to derive safety certificates.

The key idea of our CCD algorithm in Alg.\ref{alg:ccd_refinement} is to recursively check the CCD condition (Eq.~\ref{eq:ccd}) and subdivide path segments until all segments are certified collision-free. Since longer segments are more likely to violate the condition, the algorithm always inspects the longest segment first. Accordingly, all segments are stored in a priority queue $\mathcal{Q}$ sorted by $L$ (lines~\ref{ln:seg0}-\ref{ln:seg2}). At each iteration, the longest segment is extracted and evaluated for collisions (line~\ref{ln:pop}). 

If a path segment ${S}_i$ is not free from collision, the segment is bisected at its midpoint $\mathbf{p}_{i+\frac{1}{2}}$ (line~\ref{ln:bisect}) with the orientation of $\mathbf{p}_{i+\frac{1}{2}}$ interpolated from $\mathbf{p}_{i}, \mathbf{p}_{i+1}$. If $d(\mathbf{p}_{i+\frac{1}{2}}) \leq R$, a pose correction is applied to relocate $\mathbf{p}_{i+\frac{1}{2}}$ to a collision-free position, while keeping its orientation unchanged. (lines~\ref{ln:col}-\ref{ln:pose_corr}, detailed in Sec.~\ref{subsection:pose_corr}). 
However, since the corrected segment $S'_i = \{\mathbf{p}_{i}, \mathbf{p}_{i+\frac{1}{2}}, \mathbf{p}_{i+1}\}$ may not be kinematically feasible for the robot to follow (\eg due to non-holonomic constraint), $S'_i$ are adjusted to maintain consistency with the kinematic constraints (line~\ref{ln:ori}, detailed in Sec.~\ref{subsection:ori_align}).
Moreover, the original time interval $\Delta t_i$ is divided in proportion to the displacement bound of the two subdivided segments $S_i^{(1)} = \{ \mathbf{p}_{i}, \mathbf{p}_{i+\frac{1}{2}} \}$, $S_{i}^{(2)}=\{ \mathbf{p}_{i+\frac{1}{2}}, \mathbf{p}_{i+1}\}$ and $\Delta t_i = \Delta t_i^{(1)} + \Delta t_{i}^{(2)}$ (line~\ref{ln:time}). The two resulting sub-segments, together with their adjusted time intervals, $\mathcal{T}'_i = (S_i^{(1)} \cup S_{i}^{(2)}, \{\Delta t_i^{(1)}, \Delta t_{i}^{(2)}\})$ are pushed back into the queue (lines~\ref{ln:q1}-\ref{ln:q2}). 
Otherwise, the segment is considered collision-free and is therefore removed from the queue. In addition, segments whose lengths fall below a predefined threshold are exempted from further bisection to prevent excessive subdivision. As a result, safe regions are represented with sparse pose distributions, while such risky regions are adaptively refined with denser poses.

\subsection{Pose Correction} \label{subsection:pose_corr}

This section describes the pose correction strategy, which resolves collisions by shifting risky configurations to collision-free regions. Rather than relying on computationally expensive explicit geometric reasoning (e.g., evaluating Minkowski sums), our approach simplifies the collision resolution by decomposing it into an efficient two-stage process: (a) determining the SD, which maximizes the clearance from the obstacles, and (b) relocating the colliding pose along SD until it is guided into a valid attraction basin for subsequent optimization, achieving {\em maximal clearance} from surrounding obstacles. In terms of computational complexity, the pose correction step operates in $\mathcal{O}(1)$.

{First, a signed distance field $\phi(c)$ is computed on the grid map $\mathcal{M}$ for all grid cells $\forall c \in \mathcal{M}$ based on the Chamfer distance. For a robot pose $\mathbf{p} = (\mathbf{x}, \theta) \in \mathbb{R}^2 \times \mathbb{S}^1$, let $c_\mathbf{x} \in M$ be the grid cell containing $\mathbf{x}$.}

\subsubsection{Separation direction (SD) determination} \label{subsubsection:sd}
{To resolve insufficient obstacle clearance, positions with $\phi(c_\mathbf{x}) < R$ are identified as correction targets. Here, $R$ denotes a conservative threshold for clearance evaluation.
When $\mathbf{x}$ is not on the obstacle boundary (i.e., $0 < \phi(c_\mathbf{x}) < R \; \mathrm{or} \; \phi(c_\mathbf{x}) < 0$), we find $\mathbf{b} = \arg\min_{\mathbf{t} \in \partial \mathcal{O}} \|\mathbf{t} - \mathbf{x}\|$ representing the minimum distance between $\mathbf{x}$ and the obstacle boundary. Then, we set the SD parallel to $ \pm(\mathbf{x}-\mathbf{b})$. However, when $\mathbf{x}$ is on the obstacle boundary (i.e., $\phi(c_\mathbf{x}) = 0$), we find an optimal displacement $\mathbf{u}^{*}$ via a least-norm solution \cite{ben2003generalized}. By linearizing the clearance condition $\phi(\mathbf{x}+\mathbf{u}) \ge R$ and applying the Eikonal property ($\|\nabla \phi\| = 1$), the displacement $\mathbf{u}^*$ that minimizes $\|\mathbf{u}\|$ is defined as:}
\begin{align}\label{eq:u_opt} 
\mathbf{u}^{*}
&=
\left( \nabla\phi(\mathbf{x})^\top \right)^\top \left( \nabla\phi(\mathbf{x})^\top \left( \nabla\phi(\mathbf{x})^\top \right)^\top \right)^{-1} \left( R - \phi(\mathbf{x}) \right) \nonumber \\
&=
\left(R-\phi(\mathbf{x})\right)\nabla\phi(\mathbf{x}) \quad (\because \|\nabla\phi\| = 1).
\end{align}
{Consequently, the SD, $\mathbf{v}$, can be obtained from Eq.~\ref{eq:u_opt} by normalizing $\mathbf{u}^*$, which yields $\mathbf{v} = \nabla\phi(\mathbf{x})$. 

Finally, we categorize the spatial states into three scenarios based on $\phi(c_\mathbf{x})$ to find $\mathbf{v}$:}

\begin{figure}[htb!]
\centering
\begin{subfigure}[t]{0.32\columnwidth}
  \includegraphics[width=\linewidth]{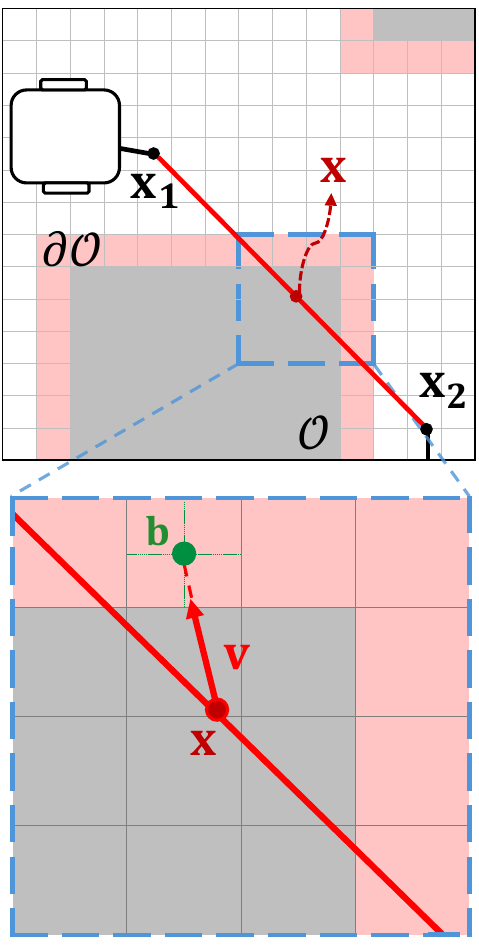}
  \caption{}\label{fig:in}
\end{subfigure}
\hfill
\begin{subfigure}[t]{0.32\columnwidth}
  \includegraphics[width=\linewidth]{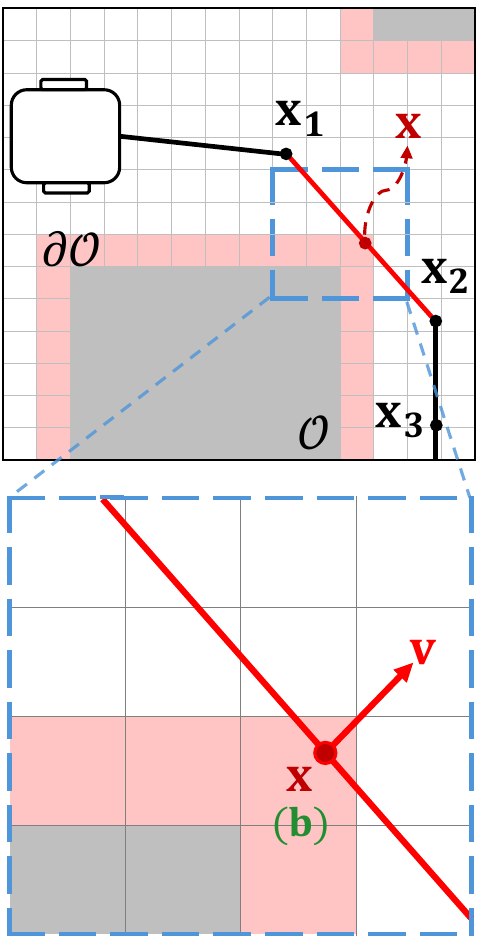}
  \caption{}\label{fig:out}
\end{subfigure}
\hfill
\begin{subfigure}[t]{0.32\columnwidth}
  \includegraphics[width=\linewidth]{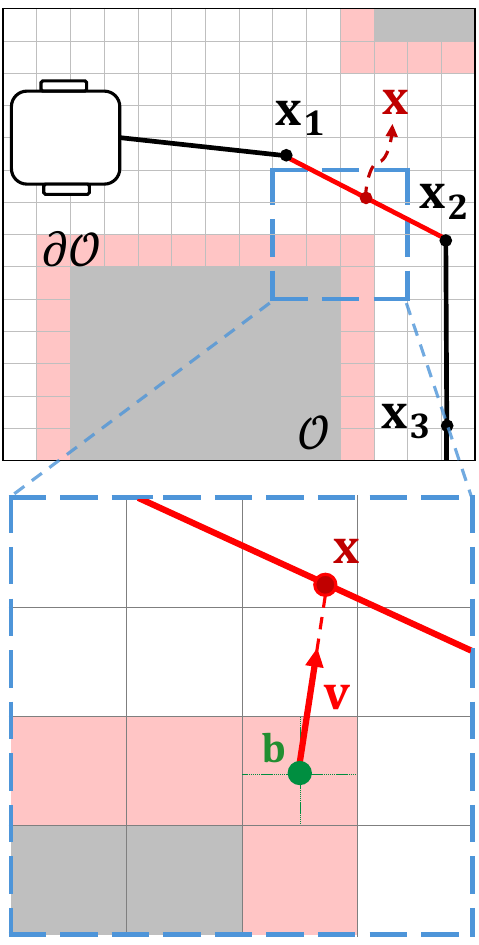}
  \caption{}\label{fig:bound}
\end{subfigure}
\caption{{\bf Separation direction $\mathbf{v}$} determined when (a) the robot's position is located at $\mathbf{x}$ inside an obstacle (gray grids), (b) on the obstacle boundary (red grids), (c) outside the obstacle but still close to the obstacle.}
\label{fig:sep_direction}
\end{figure}

\begin{enumerate}
\item[(a)]
{If $\phi(c_\mathbf{x}) < 0$, we set $\mathbf{v} = \frac{\mathbf{b} - \mathbf{x}}{\|\mathbf{b} - \mathbf{x}\|}$ as the outward retraction direction from the obstacle (Fig.~\ref{fig:sep_direction}(\subref{fig:in})).}

\item[(b)]
{If $\phi(c_\mathbf{x}) = 0$, we set $\mathbf{v} = \nabla\phi(c_\mathbf{x})$ as the direction of increasing clearance (Fig.~\ref{fig:sep_direction}(\subref{fig:out})).}

\item[(c)]
{If $0 < \phi(c_\mathbf{x}) < R$, we set $\mathbf{v} = \frac{\mathbf{x} - \mathbf{b}}{\|\mathbf{x} - \mathbf{b}\|}$ as the outward direction from the nearest boundary to satisfy the clearance bound (Fig.~\ref{fig:sep_direction}(\subref{fig:bound})).}
\end{enumerate}

\subsubsection{Pose relocation} 
After determining the separation direction $\mathbf{v}$, the pose has to be relocated to guarantee a sufficient collision clearance margin. This relocation is carried out via a line search along $\mathbf{v}$. 

First, starting from $\mathbf{x}$, a ray is cast along $\mathbf{v}$ and discretized into a set of grid cells $C$ up to the ray length $d_{\max}$ \cite{bresenham1998algorithm}. For penetrated poses, $d_{\max}$ is adaptively extended by $|\phi(c_\mathbf{x})|$ to ensure that the search extends beyond the obstacle. Then, to ensure the pose is relocated to a point of maximum safety, a directional hill-climbing procedure is applied by comparing the distances to neighboring cells in $C$ starting from $c_\mathbf{x}$. The search ends when (i) $C$ is exhausted, or (ii) no further increase in clearance is observed.
While the initial relocation extracts the pose from the obstacle, complex configuration space may prevent a point-wise correction from fully achieving sufficient clearance. To address this issue, the achieved clearance is evaluated after the initial pose relocation, and if it remains insufficient, the separation direction is recomputed at the updated location, followed by an additional relocation along the gradient direction. Since the search length is bounded by a fixed maximum in our implementation, the practical computational cost remains constant.

\subsection{Local Kinematic Feasibility Adjustment} \label{subsection:ori_align}

When the positions of poses are updated through the insertion and correction procedures described in Sec.~\ref{subsection:ccd} and \ref{subsection:pose_corr}, their orientations may not be kinematically feasible, which can prevent the robot from reaching the subsequent pose and potentially cause collisions. Since this study considers a wheeled mobile robot subject to non-holonomic kinematic constraints, such infeasibility must be addressed during trajectory refinement. {This step provides a more stable initialization by guiding the trajectory into a valid attraction basin, thereby ensuring robust convergence for the subsequent optimization stage.}

To implement this strategy, a local orientation adjustment step is performed. After $\mathbf{p}_{i}$ is refined, not only the orientation of $\mathbf{p}_{i}$ but also those of its immediate neighbors $\mathbf{p}_{i-1}, \mathbf{p}_{i+1}$ have to be reconsidered. In order to maintain kinematic feasibility, the orientations of these poses are updated such that the relative configuration conforms to the non-holonomic kinematic constraint, ensuring that the consecutive poses lie on a common arc of constant curvature \cite{rosmann2012trajectory}.

\section{Experiments}\label{sec:exp}
In this section, we present the experimental results evaluated through both simulation and real-world testing. The experiments were designed to demonstrate the effectiveness of the method in path planning within cluttered environments. In the experiments, we compared ART-TEB (ours) against two TEB-based baselines, TEB \cite{rosmann2012trajectory} and its enhanced variant egoTEB \cite{smith2020egoteb}, as well as DWA \cite{fox1997dynamic} and an MPC local planner \cite{rosmann2021online}, which are widely used as baselines.

\subsection{Simulation Experiments} \label{subsection:sim_exp}

\begin{figure}[htb!]
    \centering
    \includegraphics[width=\columnwidth]{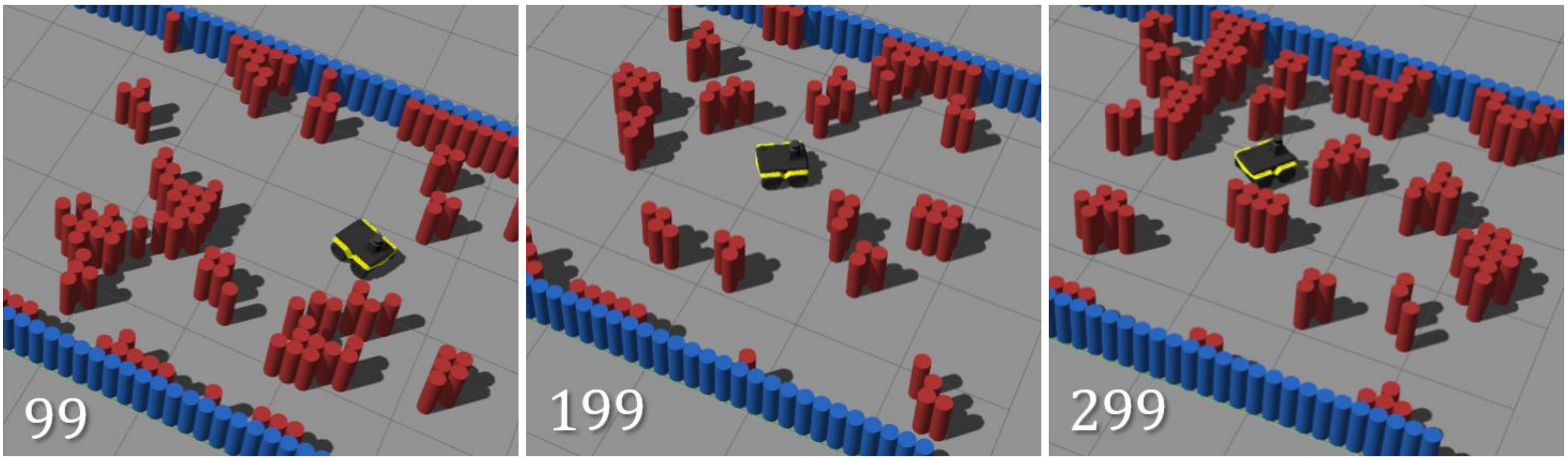}
    \caption{{\bf BARN dataset with benchmark number.} The Jackal robot navigates through 300 different environments with varying obstacle densities, represented by red cylinders.}
    \label{fig:simenv}
\end{figure}

\begin{figure*}[ht!]
{\includegraphics[width=\textwidth]{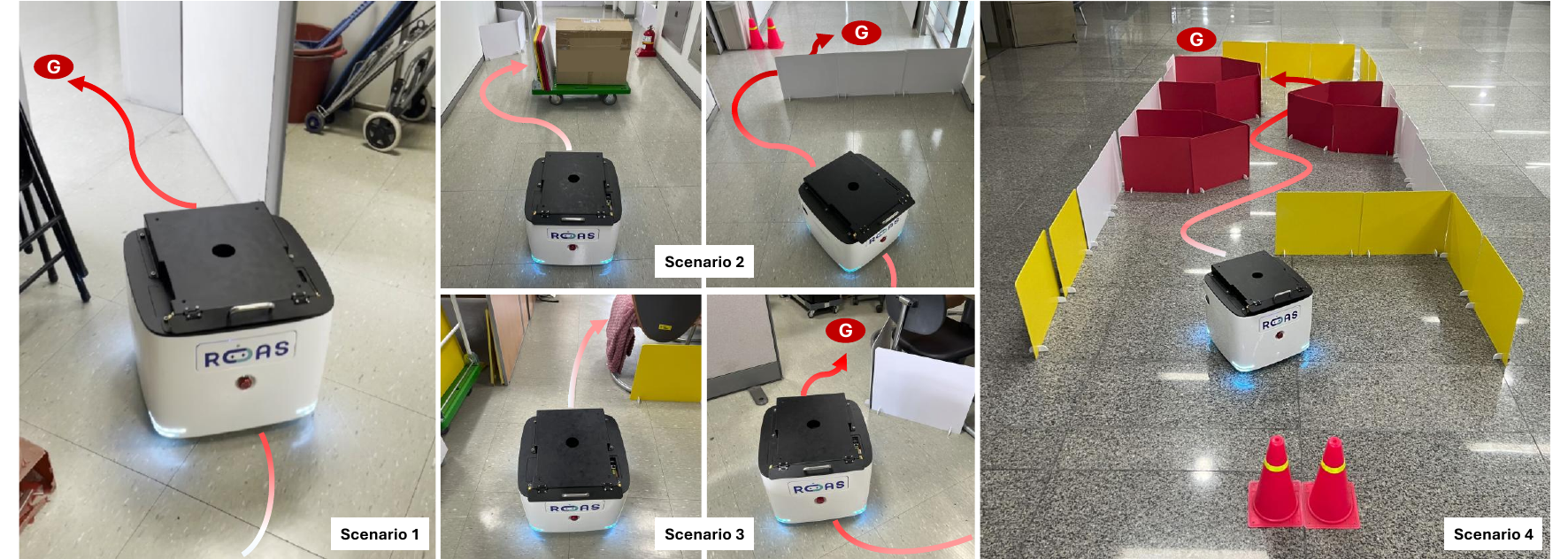}}
\captionof{figure}{ \textbf{Real-world experimental scenarios designed to evaluate performance.} Scenario 1 requires the robot to pass through a narrow doorway. Scenario 2 involves navigating past a cart and a tight passage in a confined corridor. Scenario 3 features a tight corner in an office environment that leads to the goal. Scenario 4 presents a challenging test environment with sharp turns formed by barriers. The red circle labeled “G” denotes the goal position, while the arrows illustrate the robot's navigation path.}
\label{fig:realworld}
\end{figure*}

As shown in Fig.~\ref{fig:simenv}, the simulation experiments were conducted in the Gazebo simulator using the BARN dataset \cite{perille2020benchmarking}, which provides a diverse distribution of static obstacles. All simulation experiments are conducted on an AMD Ryzen 5 3600 CPU processor running Ubuntu 20.04 with ROS Noetic. The robot used in simulation experiments is a four-wheeled differential-drive Jackal robot.

All algorithms were evaluated across 300 BARN environments, with three independent trials per environment, yielding a total of 900 runs. The start and goal positions were identical across all methods. Furthermore, we investigated the impact of the planning horizon by using the baseline setting ($3.0s$) and the $3\times$ reduction ($1.0s$).

All parameters shared by the TEB variants were set identically, except for \texttt{dt\_ref}, which determines the temporal resolution of the trajectory. To generate a coarser plan, ART-TEB used 0.6, while TEB and egoTEB used the default value (0.3). All costmap-related parameters were kept identical across all methods.

\begin{table}[htb]
\renewcommand{\arraystretch}{1.2}
\setlength{\tabcolsep}{3pt}
\centering
\begin{tabular}{lc|c|c|c}
\toprule[1.5pt]
\begin{tabular}[c]{@{}c@{}}Method\end{tabular} & \begin{tabular}[c]{@{}c@{}}Success \\ Rate (\%)\end{tabular} & \begin{tabular}[c]{@{}c@{}}Avg. Planning \\ Time (ms)\end{tabular} & \begin{tabular}[c]{@{}c@{}}Max. Planning \\ Time (ms)\end{tabular} & \begin{tabular}[c]{@{}c@{}}Traversal \\ Time (s)\end{tabular}\\
\midrule
\makecell[c]{DWA}  & 54.67 & 11.76 & 14.69 & 30.76 \\\addlinespace[1.5pt]
\makecell[c]{MPC}   & 31.78 & 53.65 & 153.91 & 36.95 \\\addlinespace[3pt]
\makecell[c]{TEB}   & 58.75 / 83.78 & 5.27 / 8.99 & 12.78 / 19.60 & 17.68 / 16.79 \\
\addlinespace[3pt]
\makecell[c]{egoTEB} & 86.83 / 60.67 & 3.25 / 4.73 &  7.33 / 17.51 & \textbf{15.75} / 15.18\\
\addlinespace[3pt]
\makecell[c]{ART-TEB}  & \textbf{99.25} / \textbf{95.11} & 2.26 / 2.37  & \textbf{5.15} / 5.88 & 15.93 / 15.69\\
\addlinespace[3pt]
\makecell[c]{ART-TEB\\w/o CCD}  & 82.56 / 52.33 & \textbf{1.97} / \textbf{1.98}  & 5.18 / \textbf{4.79} & 16.17 / \textbf{15.1}\\
\bottomrule[1.5pt]
\end{tabular}
\caption{\textbf{Quantitative performance comparison in simulation.} Avg. Planning Time is averaged over all planning steps, whereas Max. Planning Time is obtained by averaging the maximum planning time per trial. Traversal Time denotes the time required for the robot to travel from start to goal. For TEB variants, -/- in each cell indicates results corresponding to planning horizons of $1.0 s$ and $3.0 s$, respectively.}
\label{tab:sim-results}
\end{table}

\textbf{Comparison} As summarized in Table~\ref{tab:sim-results}, ART-TEB consistently outperformed the baselines in both success rate and planning times. In terms of success rate, our method achieved 99.25\% at a horizon of $1.0s$ and maintained 95.11\% at a horizon of $3.0s$. This corresponds to a success rate up to $1.69\times$ higher than the TEB variants across the two horizons. In addition, our method achieved a success rate up to $1.82\times$ higher than DWA and up to $3.12\times$ higher than MPC.
In terms of computational efficiency, our method achieved significantly faster planning times, with up to $3.79\times$ speedup over the TEB variants. Our planner was also $5.2\times$ faster than DWA and nearly $23.7\times$ faster than MPC. Unlike the other TEB variant approaches, the efficiency of our method mainly 
comes from the adaptive pose distribution, which reduces the 
computational burden. 
Notably, while the average planning time of TEB increased by $1.7\times$ and egoTEB by $1.45\times$ as the horizon grew from $1.0s$ to $3.0s$, our method exhibited only a marginal increase of $1.05\times$. This indicates that the planning time of our approach is minimally affected by the planning horizon, thereby highlighting its efficiency.
Finally, in terms of maximum planning time, our method reduced it by $2.48\times/3.33\times$ compared to TEB and by $1.42\times/2.98\times$ compared to egoTEB, and further reduced it by approximately $29\times$ compared to MPC. These results demonstrate that our method mitigates worst-case planning delays, ensuring reliable performance in real-time operations. 

\begin{table*}[htb]
\renewcommand{\arraystretch}{1.0}
\setlength{\tabcolsep}{3pt}      
\centering
\begin{tabular}{lccc|ccc|ccc|ccc|ccc}
\toprule[1.5pt]
\multirow{2}{*}{Scenario} 
  & \multicolumn{3}{c}{Task Success} 
  & \multicolumn{3}{c}{Progress Rate (\%)} 
  & \multicolumn{3}{c}{Avg. Planning Time (ms)} 
  & \multicolumn{3}{c}{Max Planning Time (ms)}
  & \multicolumn{3}{c}{Avg. Traversal Time (s)}\\
\cmidrule(lr){2-4}\cmidrule(lr){5-7}\cmidrule(lr){8-10}\cmidrule(lr){11-13}\cmidrule(lr){14-16}
  & TEB & egoTEB & ART-TEB
  & TEB & egoTEB & ART-TEB 
  & TEB & egoTEB & ART-TEB
  & TEB & egoTEB & ART-TEB
  & TEB & egoTEB & ART-TEB \\
\midrule
Scen. 1    & 0/5 & 1/5 & \textbf{5/5}  & 41.77 & 47.76 & \textbf{100.0} & - & 3.81 & \textbf{3.06} & - & 13.61 & \textbf{8.96} & - & 10.03 & \textbf{10.03} \\
Scen. 2  & 0/5 & 0/5 & \textbf{5/5}  & 49.58 &  49.5 & \textbf{100.0} & - & - & \textbf{3.87} & - & - & \textbf{10.09} & - & - & \textbf{20.15} \\
Scen. 3    & 0/5 & 0/5 & \textbf{4/5}  & 36.27 & 37.85 & \textbf{95.49} & - & - & \textbf{6.71} & - & - & \textbf{10.58} & - & - & \textbf{18.94} \\
Scen. 4  & 0/5 & 0/5 & \textbf{5/5}  & 14.97 & 51.81 & \textbf{100.0} & - & - & \textbf{5.78} & - & - & \textbf{12.94} & - & - & \textbf{22.88}\\
\bottomrule[1.5pt]
\end{tabular}
\caption{\textbf{ Performance in real-world environments.} ART-TEB achieved the highest task success across all scenarios, with lower planning times compared to the baselines. “$-$” denotes cases in which the method failed to succeed.}
\label{tab:real-results}
\end{table*}


\textbf{Ablation Study}
We conducted an ablation study on the segment-wise CCD (described in Sec.~\ref {subsection:ccd}). An ablation study of the pose correction step (described in Sec.~\ref {subsection:pose_corr}) is not feasible, as the pose remains in collision with obstacles, leading to excessive (potentially infinite) subdivision. 

Removing CCD reduces the success rate by almost half ($55\%$) under the longer-horizon setting, demonstrating that the CCD algorithm plays a critical role in maintaining collision-free trajectories. In terms of computational cost, the results show a slight reduction in planning time compared to the proposed method. This indicates that the additional complexity introduced by CCD remains relatively low.

\subsection{Real-world Experiments}

The real-world experiments were conducted using a differential-drive mobile robot equipped with 2D LiDAR-based SLAM \cite{macenski2021slam}. It is powered by an Intel i5 processor with 8 GB of RAM, running on Ubuntu 18.04 with ROS Melodic, allowing on-board execution of navigation tasks. Due to poor performance in simulation experiments, we excluded DWA and MPC from the real-world experiments.

We deployed the system in various environments, as shown in Fig.~\ref{fig:realworld}. These environments include a narrow doorway, a cluttered environment with tight passages and corners, and a challenging test environment. Each method was evaluated five times in each scenario. Like Sec.~\ref{subsection:sim_exp}, we measured success rate, traversal time, average, and maximum planning time. For failed runs, we additionally measured the progress rate, defined as the percentage of the global plan completed before termination.
The following presents a comparative analysis across four real-world scenarios : 

\begin{enumerate}

\item \textbf{Scenario 1} As shown in Table~\ref{tab:real-results}, ART-TEB completed all trials, while egoTEB succeeded once and TEB failed in all attempts. Both baselines often failed at the doorway. Planning time was $1.25\times$ faster than egoTEB, even in real-world settings.

\item \textbf{Scenario 2} Our method achieved 100\% success across all trials, while both TEB and egoTEB failed entirely. TEB and egoTEB collided with the cart, doors, or obstacles surrounding the narrow gaps. These failures highlight the sparsity issues and obstacle penetration tendencies of the trajectory inherent to the TEB process, as discussed in Sec.~\ref {section:intro}, which become more pronounced in real-world environments. By contrast, our method overcame these challenges and exhibited robustness to such complexities.

\item \textbf{Scenario 3} Similar to Scenario 2, both TEB and egoTEB failed in all trials, while our method succeeded in all but one. Scenario 3 was the most cluttered of the four, featuring dense, irregular obstacles. This complexity likely caused the single failure of our method; nevertheless, it still achieved a higher success rate than the baselines.

\item \textbf{Scenario 4} This scenario features a challenging environment composed of consecutive corners. While TEB and egoTEB failed in all trials, our algorithm successfully navigated the environment.

\end{enumerate}

\begin{figure}[htb!]
    \centering
    \includegraphics[width=\columnwidth]{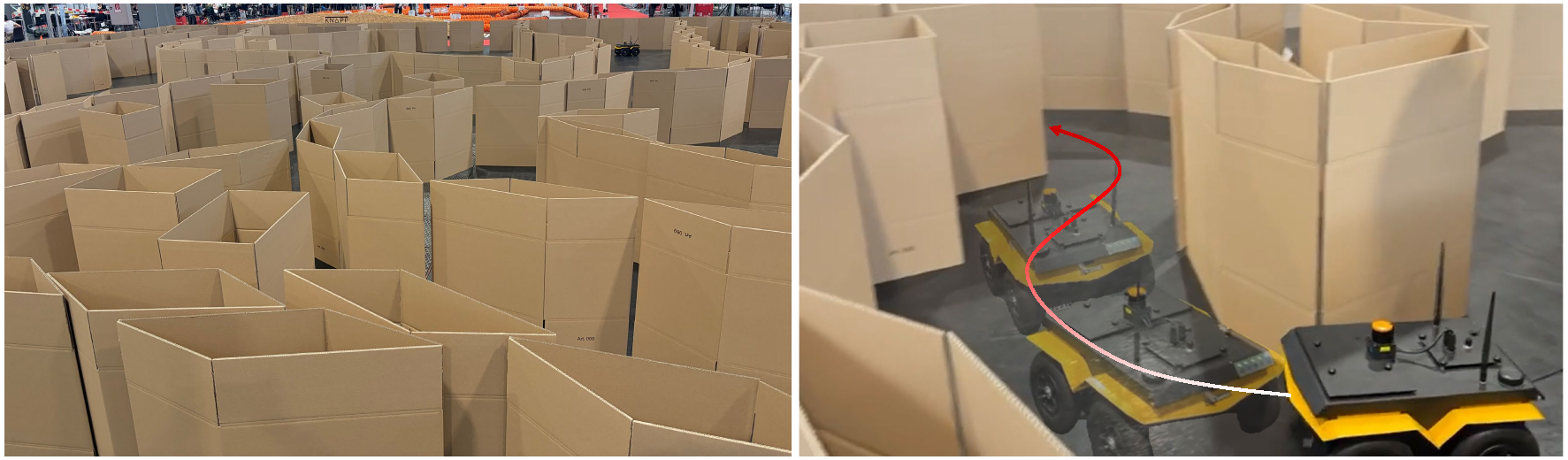}
    \caption{{\bf Real-world BARN Challenge environment.} The benchmark features narrow, cluttered passages constructed with cardboard boxes.}
    \label{fig:barn}
\end{figure} 

{In addition to the aforementioned comparative analyses, the overall capability and practical robustness of our method were comprehensively validated in the {\em 2026 BARN Challenge}. As shown in Fig.~\ref{fig:barn}, this benchmark provides a challenging real-world environment. Competing in both the simulation and real-world phases, our method achieved second place overall, successfully outperforming the previous year's winning teams that represent recent state-of-the-art local planning approaches \cite{lu2025decremental,mohammad2024gp,mohammad2025soft}. This achievement proves the effectiveness of ART-TEB for highly challenging, real-world deployments.}

\section{Conclusion}
This paper introduced ART-TEB, a novel planner that enables collision-free trajectory generation in challenging scenarios such as narrow passages. We evaluated the effectiveness of our planner through simulation and real-world experiments across various scenarios, comparing it to state-of-the-art approaches. Our current algorithm lacks an explicit motion prediction module, limiting its ability to effectively avoid highly dynamic obstacles. Future work will focus on integrating a motion prediction component to improve performance in dynamic environments.






\bibliographystyle{IEEEtran}
\bibliography{main}

\end{document}